\documentclass[10pt,twocolumn,letterpaper]{article}

\usepackage{cvpr}
\usepackage{times}
\usepackage{epsfig}
\usepackage{graphicx}
\usepackage{amsmath}
\usepackage{amssymb}
\usepackage{mathtools}

\newcommand{\SOTA}{state-of-the-art }

\usepackage{mwe}
\usepackage[markcase=noupper
]{scrlayer-scrpage}
\ohead{}
\usepackage[pagebackref=true,breaklinks=true,letterpaper=true,colorlinks,bookmarks=false]{hyperref}

\cvprfinalcopy 

\def\cvprPaperID{9} 

\begin{document}

\title{Empirical Perspectives on One-Shot Semi-supervised Learning}

\author{Leslie N. Smith\\
US Naval Research Laboratory\\
Washington, DC\\
{\tt\small leslie.smith@nrl.navy.mil}
\and
Adam Conovaloff\\
NRC Postdoctoral Fellow\\
US Naval Research Laboratory\\
{\tt\small adam.conovaloff@nrl.navy.mil}
}

\maketitle

\begin{abstract}
One of the greatest obstacles in the adoption of deep neural networks for new applications is that training the network typically requires a large number of manually labeled training samples.
We empirically investigate the scenario where one has access to large amounts of unlabeled data but require labeling only a single prototypical sample per class in order to train a deep network (i.e., one-shot semi-supervised learning).
Specifically, we investigate the recent results reported in FixMatch \cite{sohn2020fixmatch} for one-shot semi-supervised learning to understand the factors that affect and impede high accuracies and reliability for Cifar-10.
For example, we discover that one barrier to one-shot semi-supervised learning for image classification is the class accuracy unevenness of the training.
These results point to solutions that might enable more widespread adoption of one-shot semi-supervised training methods for new applications.
\end{abstract}

\section{Introduction}

Deep learning has proven to be highly successful in image classification tasks in recent years.
One of the greatest barriers in the adoption of deep neural networks for new applications is that training \SOTA deep networks typically require require hundreds or thousands of labeled samples per class to perform at high levels of accuracy.
Unfortunately, manual labeling is labor intensive, time consuming, and tedious.
One option is to use Mechanical Turk or a similar service but this is not possible if the data requires expertise or the data is confidential.

In the common real-world scenario where one has access to large amounts of unlabeled data and labeling is labor intensive, we suggest that an ideal solution would be able to train deep neural networks from this data pool by manually labeling as little as one iconic image per class and resulting in networks trained to performance levels that are comparable to the performance with fully supervised trained networks.
Such a scenario eliminates the burden of manually labeling training data and eliminates the time consuming aspects of labeling as an obstacle for new applications of image classification, especially in fields where labeling requires an expert's time and effort, such as in medical, defense, and scientific applications.

In addition to these applied motivations to study one-shot semi-supervised learning, one-shot semi-supervised learning is scientifically useful to study because of its especially high sensitivity to all the factors affecting training, such as hyper-parameter settings.
Hence, one can observe phenomena that are masked in fully supervised learning with large numbers of labeled data case in order to more easily find optimal training settings.

In this paper we experimentally investigate the one-shot semi-supervised scenario to better understand the factors that affect the training and test performance.
We build on FixMatch \cite{sohn2020fixmatch}, which is (to the best of our knowledge) the first method to demonstrate reasonable test accuracies for Cifar-10 while starting with only one example per class.
That is, the authors report test accuracies between the range of 48.58\% and 85.32\% using four choices for the one labeled sample per class, which is a remarkable result in spite of the high variability of the performance.
We investigated these experiments and we discovered that one significant factor that causes this variability and prevents higher test results is a class imbalance problem.  

Specifically, we show in this paper that one-shot semi-supervised training based on FixMatch generally creates a network that works with near perfect accuracy for some classes and near zero accuracy on some of the other classes.
These results are important because understanding the problems paves the way to overcome the obstacles to accurate and reliable one-shot semi-supervised training. 

%

Our contributions are:
\begin{enumerate}
	\item We investigate one-shot semi-supervised learning to determine the factors that reduce performance and increase variability.
	\item We demonstrate on Cifar-10 that some classes are learned to near 100\% test accuracy but other classes have near 0\% accuracy.
	\item We discuss how these results point to potential solutions.
\end{enumerate}

\begin{table*}
	\begin{center}
		\begin{tabular}{|c|c|c|c|c|c|c|c|c|c|c|}
			\hline
			All & plane & auto & bird & cat & deer & dog & frog & horse & ship & truck \\
			\hline\hline
			$63 \pm 5$  & $17 \pm 32$  & $98.2 \pm .5$ & $21 \pm 11$ & $67 \pm 45$ & $96 \pm 3$ & $0 \pm 0$ & $25 \pm 48$ & $91 \pm 11$ & $98.3 \pm .6$ & $96 \pm 1.4$  \\
			
			$ 71 \pm 5$  & $ 96 \pm 2$  & $97.6 \pm .8$ & $60 \pm 7$ & $.05 \pm .06$ & $25 \pm 50$ & $24 \pm 48$ & $97 \pm 2.6$ & $93 \pm 13$ & $72 \pm 48$ & $98 \pm .1$  \\
			
			$ 65 \pm 2$  & $ 79 \pm 34$  & $98 \pm 0.2$ & $22 \pm 6$ & $52 \pm 50$ & $98 \pm 1$ & $0 \pm 0$ & $75 \pm 45$ & $15 \pm 10$ & $73 \pm 49$ & $97.4 \pm .4$  \\
			
			$ 64 \pm 5$  & $ 96 \pm 1$  & $4.8 \pm 0.2$ & $66 \pm 2$ & $23 \pm 47$ & $97 \pm 1$ & $48 \pm 54$ & $97 \pm 1$ & $10 \pm 12$ & $74 \pm 46$ & $74 \pm 47$  \\
			
			$ 45 \pm 4$  & $ 19 \pm 32$  & $74 \pm 49$ & $12 \pm 5$ & $8 \pm 8$ & $0 \pm 0$ & $25 \pm 49$ & $99 \pm 1$ & $0.2 \pm 0.2$ & $97 \pm 0.8$ & $97 \pm 1$  \\
			
			$ 58 \pm 6$  & $ 32 \pm 43$  & $74 \pm 49$ & $9 \pm 4$ & $0.1 \pm 0.1$ & $0.2 \pm 0.2$ & $7 \pm 5$ & $96 \pm 0.8$ & $94 \pm 11$ & $72 \pm 46$ & $96 \pm 2$  \\
			
			\hline
		\end{tabular}
	\end{center}
	\caption{One shot semi-supervised test accuracies on Cifar-10 for 6 different data sample seeding (i.e., choice of which image to label). Each row shows the results for a single seed. Each result is the average and standard deviation of four runs. There is a significant variation in class accuracies between the choices for which images to label. The test class accuracies have a great deal more variation than the accuracy variation over all of the classes. }
	\label{tab:classAcc4}
\end{table*}

\begin{table*}
	\begin{center}
		\begin{tabular}{|c|c|c|c|c|c|c|c|c|c|}
			\hline
			plane & auto & bird & cat & deer & dog & frog & horse & ship & truck \\
			\hline\hline
			$98.2 $  & $ 4.7 $  & $ 66.3 $  & $ 0 $  & $ 96 $  & $ 94.8 $  & $ 96.6 $  & $ 19.2 $  & $ 4.3 $  & $ 2.6 $ \\
			$ 95.9  $  & $ 4.7  $  & $ 62.7  $  & $ 0  $  & $ 95.5  $  & $ 1.9  $  & $ 95.4  $  & $ 20.6  $  & $ 95.2  $  & $ 96.4 $ \\
			$ 96.2  $  & $ 4.6  $  & $ 66.4  $  & $ 0  $  & $ 97.3  $  & $ 94.3  $  & $ 98.3 $  & $ 0  $  & $ 98.1 $  & $ 98 $ \\
			$ 96.3  $  & $ 5  $  & $ 67.3  $  & $ 93.2  $  & $ 97.7  $  & $ 0  $  & $ 98.2  $  & $ 0  $  & $ 97.2  $  & $ 97.5 $ \\
			\hline
		\end{tabular}
	\end{center}
	\caption{One shot semi-supervised test accuracies for all four runs contained in the fourth row in Table \ref{tab:classAcc4} (i.e., seed 3).  Even with the same images labeled, there is a great deal of variation in the class results.}
	\label{tab:classAcc1}
\end{table*}

\section{Related Work}

While our work primarily builds on FixMatch \cite{sohn2020fixmatch}, it is related to other fields of research, including unsupervised self-training, semi-supervised learning, noisy label robustness, class imbalance, and data augmentation.  A thorough review of this literature is beyond the scope of this paper so in this Section we briefly mention only the most relevant papers.

Our scenario superficially bears similarity to few-shot meta learning \cite{koch2015siamese,vinyals2016matching,finn2017model,snell2017prototypical}, which is a highly active area of research.
The majority of the work in this area relies on a large labeled dataset with similar data statistics but this can be an onerous requirement for new applications.
While there is some recent efforts in unsupervised pretraining for few-shot meta learning \cite{hsu2018unsupervised,antoniou2019assume}, our experiments with these methods demonstrated their inability to adequately perform in one shot learning to reasonable accuracy levels.
Specifically, one-shot learning with only five classes obtained a test accuracy of less than 40\% and the accuracy dropped sharply when increasing the number of classes.

The potential success of one-shot semi-supervised learning relies on the observation that neural networks are relatively robust to label noise \cite{algan2019image}.  
Hence, if at any stage a small percent (less than 20\%) of the high confidence labels are wrong, the impact on the future performance is minimal.
The intuition here is that noisy labels benefits from the class imbalance problem \cite{johnson2019survey}; that is, the majority (of correctly labeled samples) overwhelms the minority (incorrectly labeled samples).



\section{FixMatch}

FixMatch \cite{sohn2020fixmatch} is based on a combination of pseudo-labeling and consistency regularization, two methods for semi-supervised learning.
Pseudo-labeling uses inference on the unlabeled data and weakly augmented versions of the data generates pseudo-labels (i.e., an artificial label) and these pseudo-labels are retained only if the model produces high confidence predictions.

Consistency regularization in semi-supervised learning relies on the model producing the same classification with both the original unlabeled image as with a modified version of the image.
In FixMatch, strongly augmented versions of the data is used to train the model to produce the same pseudo-labels as produced by inference on weakly augmented versions.

FixMatch uses two forms of data augmentation: weak and strong.  
Weak augmentation consists only of random horizontal flip and pixel shifting by randomly translating the image.
For strong augmentation, FixMatch uses CutOut \cite{devries2017improved}, CTAugment \cite{berthelot2019remixmatch}, and RandAugment \cite{cubuk2019randaugment} to produce strongly augmented versions of the image.
That is, it uses CutOut followed by either of the two strong augmentation methods.
For more detailed information, please refer to the FixMatch paper \cite{sohn2020fixmatch}.

FixMatch utilizes an insight also present in other works, such as Xie, \etal \cite{xie2019self}, where they observe that noise (i.e., strong data augmentation) should be minimized when inferring pseudo-labels for the unlabeled data but should be employed during training.
We too found this to be true in our own experiments.

\section{Experiments}
\label{sec:exp}

Our experiments extend the one shot experiments in FixMatch \cite{sohn2020fixmatch}.  Specifically, we use the provided codebase\footnote{\url{https://github.com/google-research/fixmatch}} and perform experiments with Cifar-10 \cite{krizhevsky2009learning}.  
The hyper-parameter settings are identical for all of our runs and match the settings provided in the paper.

Table \ref{tab:classAcc4} shows the average test accuracy and standard deviation over 4 runs using each of the six seeds (i.e., each row is a seed or specific choice for which image to label).  The Table includes the overall accuracy over all the classes and also breaks out the individual class accuracies.
There are several observations that can be made from this Table.
First, the choice of which image to label matters.  
There are significant differences in class performance for each of the six seeds (\ie rows).
For example, the deer class accuracies are in the upper 90\% for some labeling choice and close to 0\% for others.
Second, the class variance can be much larger than the variance over all the classes.
That is, even with the same data and labeling, there can be a great deal of variation in which classes perform well or poorly but that the average over all of the classes does not vary as much.

Table \ref{tab:classAcc1}  shows this in greater detail.  It contains the class accurracies for all four runs with the seed 3 data (i.e., the fourth row of Table \ref{tab:classAcc4}).
From this Table we can see that airplane, auto, bird, deer, and frog show a small amount of change over the four runs but cat, dog, horse, ship, and truck shows a larger variation.  
For example, in the first run the cat accuracy is 0\% and the dog accuracy is 94.8\% but in the last run this switches as the cat accuracy is 93.2\% and the dog accuracy is 0\%.
This implies that in the first case the cats and dogs are being classified as dogs but in the second case the cats and dogs are being classified as dogs.

\begin{figure}[t]
	\begin{center}
		\includegraphics[width=0.95\linewidth]{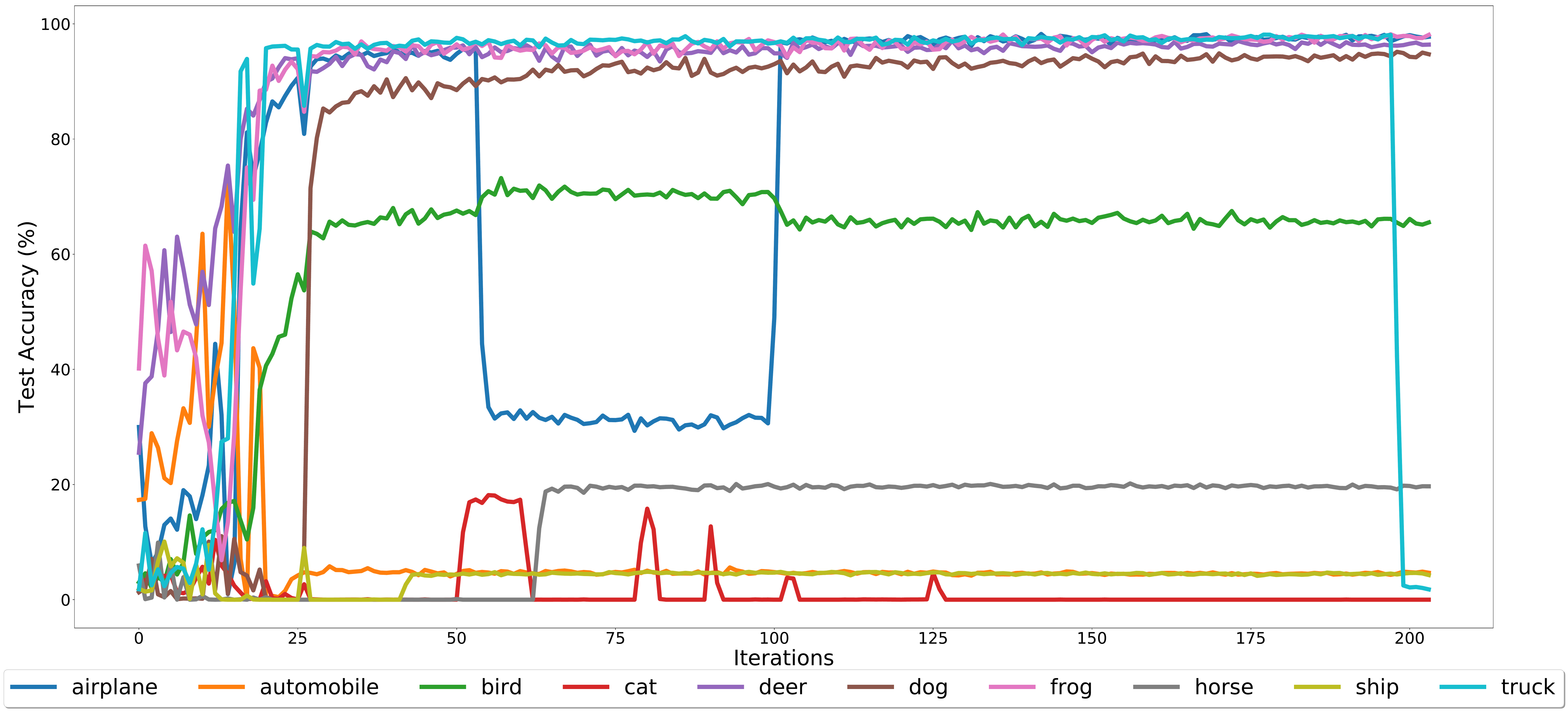}
	\end{center}
	\caption{Class test accuracies on Cifar-10 during training with seed=3 (i.e., corresponds to the results displayed in the first row of Table \ref{tab:classAcc1}).}
	\label{fig:ClassTestAccSeed3_0}
\end{figure}

\begin{figure}[t]
	\begin{center}
		\includegraphics[width=0.95\linewidth]{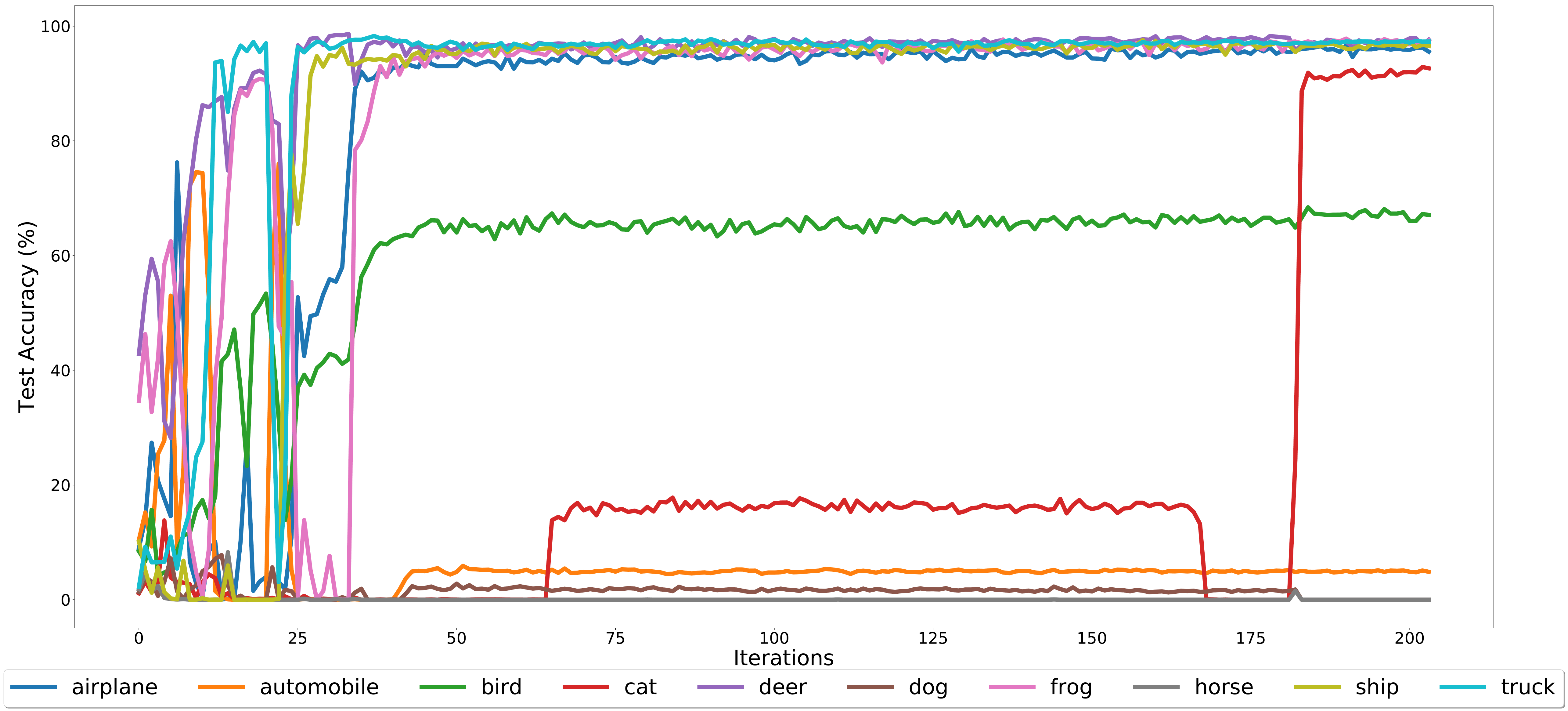}
	\end{center}
	\caption{Another example of class test accuracies on Cifar-10 with seed=3 (i.e., corresponds to the results displayed in the last row of Table \ref{tab:classAcc1}).}
	\label{fig:ClassTestAccSeed3_3}
\end{figure}

Figures \ref{fig:ClassTestAccSeed3_0} and \ref{fig:ClassTestAccSeed3_3} illustrate how the test class accuracies change during the course of the training.
The Figures are plots of the class test accuracies the same data (i.e., seed=3) but two different runs.
Figure \ref{fig:ClassTestAccSeed3_0} corresponds to the results displayed in the first row of Table \ref{tab:classAcc1} and Figure \ref{fig:ClassTestAccSeed3_3} corresponds to the results displayed in the last row.
A major feature apparent in these Figures is the dramatic difference in performance of different classes.
Another feature is the behavior during training.
That is, some classes relatively quickly learn to achieve near their top performance and remain there throughout the rest of the training but other classes demonstrate significant swings during the training.
Furthermore, which classes are in each category changes in different runs, even though the images that were labeled does not change.

\section{Discussion}

The experiments in Section \ref{sec:exp} demonstrate novel phenomena for one-shot semi-supervised learning that we analyze in this Section.  The motivation of this Section is to understand the problems in order to determine solutions that make one-shot semi-supervised learning more reliable.

Table \ref{tab:classAcc4} demonstrates great variability of the results based on the choice of which training sample is labeled.
The FixMatch paper also noted this and demonstrated the importance of labeling iconic images.
Specifically, they contructed eight datasets with varying degree of "prototypicality", which means they measure how representative the example is of its class.
The authors found that training on the most prototypical examples reached a median test accuracy of 78\% and much lower variability than shown throughout Table \ref{tab:classAcc4}.
They determined prototypicality by using fully supervised trained models, which the authors admit is impractical for real world semi-supervised training, but we argue that their results imply that it is important for practitioners to review their unlabeled data and visibly choose iconic prototypes to represent each of their classes.
This does not impose an undue hardship on the practitioner but can significantly help the algorithm.
We plan to test this as future work.

Table \ref{tab:classAcc1} shows dramatic differences between class performances, even when using the same labeled and unlabeled data.
Furthermore, Figures \ref{fig:ClassTestAccSeed3_0} and \ref{fig:ClassTestAccSeed3_3} also show dramatic differences between class performances during training.
This implies that the most benefit will come from improving the performance of the least performing classes, provided the changes do not unduly sacrifice the high performing classes.

We believe this observed difference between class performances is analogous to the class imbalance problem, which suggests utilizing techniques from that field \cite{johnson2019survey}.
Deep networks do not learn every class at the same speed or ease, even in the fully supervised case with equal numbers of labeled samples per class.
Both data balancing methods (\ie over-sampling and under-sampling) and algorithm level methods (\ie focal loss \cite{lin2017focal}) can be dynamically employed during training to provide ``affirmative action'' for the struggling classes.

For example, during training one can easily count the number of inferred samples per class when the model is used on the unlabeled data. 
The imbalance in the number per class can be used to weight the under-represented classes and compensate or overcompensate for this imbalance in order to pull the learning back to a greater equality for all the classes.
We plan to investigate these methods in one-shot semi-supervised learning as future work.

\section{Conclusions}

In this paper, we empirically investigate one-shot semi-supervised learning in order to determine the difficulties and point to potential ways to make it more accurate and reliable.
We show that the choice of labeling an iconic prototype for each class makes a significant difference in performance.

In addition, we demonstrate significantly more variation in class accuracies than is masked by only observing the overall accuracies.
We propose an analogy to class imbalance and that these problems can be mitigated by utilizing methods from the class imbalance research.

We believe that this paper provides an important step to make one-shot semi-supervised learning accurate and reliable in scenarios where one has access to large amounts of unlabeled data and labeling is labor intensive.
That is, the importance to practitioners of eliminating the requirement of substantial labeling provides strong motivation to understand the challenges and find solutions.

{\small
\bibliographystyle{ieee_fullname}
\bibliography{barrier}
}

\end{document}